\documentclass[conference,a4paper]{IEEEtran}
\IEEEoverridecommandlockouts
\usepackage{cite}
\usepackage{amsmath,amssymb,amsfonts}
\usepackage{algorithmic}
\usepackage{graphicx}
\usepackage{textcomp}
\usepackage{xcolor}
\usepackage{multirow}
\usepackage{caption}
\def\BibTeX{{\rm B\kern-.05em{\sc i\kern-.025em b}\kern-.08em
    T\kern-.1667em\lower.7ex\hbox{E}\kern-.125emX}}

\begin{document}

\title{Anonymous Jamming Detection in 5G with Bayesian Network Model Based Inference Analysis\\

\thanks{This work was funded by Deloitte \& Touche LLP's
Cyber 5G Strategic Growth Offering.}
}

\author{\IEEEauthorblockN{Ying Wang}
\IEEEauthorblockA{
\textit{Stevens Institute of Technology}\\
Hoboken, NJ, USA \\
\textit{Virginia Tech, Blacksburg, VA, USA}  \\
ywang6@stevens.edu}
\and
\IEEEauthorblockN{Shashank Jere}
\IEEEauthorblockA{\textit{Bradley Department of ECE} \\
\textit{Virginia Tech}\\
Blacksburg, VA, USA \\
shashankjere@vt.edu}
\and
\IEEEauthorblockN{Soumya Banerjee}
\IEEEauthorblockA{\textit{Virginia Modeling Analysis and Simulation} \\
\textit{Center} \\
\textit{Old Dominion University, Norfolk, VA, USA} \\
s1banerj@odu.edu}\\ 
\and
\IEEEauthorblockN{Lingjia Liu}
\IEEEauthorblockA{\textit{Bradley Department of ECE} \\
\textit{Virginia Tech}\\
Blacksburg, VA, US \\
ljliu@vt.edu}
\and
\IEEEauthorblockN{Sachin Shetty}
\IEEEauthorblockA{\textit{Virginia Modeling Analysis and}\\
\textit{Simulation Center} \\
\textit{Old Dominion University, Norfolk, VA, USA} \\
sshetty@odu.edu}
\and
\IEEEauthorblockN{Shehadi Dayekh}
\IEEEauthorblockA{\textit{Cyber 5G Strategic Growth Offering } \\
\textit{Deloitte \& Touche LLP}\\
Dallas, TX, USA \\
sdayekh@deloitte.com}
}

\maketitle

\begin{abstract}
Jamming and intrusion detection are some of the most important research domains in 5G that aim to maintain use-case reliability, prevent degradation of user experience, and avoid severe infrastructure failure or denial of service in mission-critical applications. This paper introduces an anonymous jamming detection model for 5G and beyond based on critical signal parameters collected from the radio access and core network's protocol stacks on a 5G testbed. The introduced system leverages both supervised and unsupervised learning to detect jamming with high-accuracy in real time, and allows for robust detection of unknown jamming types. Based on the given types of jamming, supervised instantaneous detection models reach an Area Under the Curve (AUC) within a range of $0.964$ to $1$ as compared to temporal-based long short-term memory (LSTM) models that reach AUC within a range of $0.923$ to $1$. 
The need for data annotation effort and the required knowledge of a vocabulary of known jamming limits the usage of the introduced supervised learning-based approach. 
To mitigate this issue, an unsupervised auto-encoder-based anomaly detection is also presented.  
The introduced unsupervised approach has an AUC of $0.987$ with training samples collected without any jamming or interference and shows resistance to adversarial training samples within certain percentage.
To retain transparency and allow domain knowledge injection, a Bayesian network model based causation analysis is further introduced.

\end{abstract}


\section{Introduction}
5G communications and its application verticals are prone to jamming due to the inherent nature of wireless radio frequency transmission. As federal agencies and businesses rely more on 5G and beyond infrastructure, they are becoming increasingly more vulnerable to sophisticated cyber attacks that may cause communication disruption which can be costly. Such jamming attacks pose severe risks to verticals, including self-driving cars, smart cities, public safety, and healthcare. Accurate and fast detection techniques of known and unknown jamming over the radio interface is the first step toward preventing such cyber threats and efficiently protecting critical communication. 

The anticipation and prevention types of defense (APD) and anomaly-based intrusion detection systems (IDS) represent the two ends in the spectrum of anti-jamming techniques. APD relies on the prior probability and knowledge of possible attacks and malfunctions of a system. In contrast, IDS focuses on learning the characteristics of the expected system response and detects anomalous cases. Preventive security measures are ineffective against unforeseen or zero-day attacks~ \cite{Linkov2013ResilienceSystems}. Cyber risk system assessments that solely depend on APD heuristics such as sums of vulnerability scores or quantities of missing patches, open ports, etc., and such metrics are widely seen as weak and potentially misleading ~\cite{Linkov2013ResilienceSystems, JansenNISTIRResearch}. 

In contrast with APD, several in-band IDS based anomaly detection techniques have been researched~\cite{Krayani2020Self-LearningCognitive-UAV-Radios}, \cite{Walton2017UnsupervisedTransmissions}, \cite{Tandiya2018DeepNetworks}, \cite{Rajendran2019UnsupervisedFeatures}. Authors in \cite{Walton2017UnsupervisedTransmissions} proposed an unsupervised anomaly detection method based on a combination of Long-Short Term Memory (LSTM) and Mixture Density Network (MDN) applied to temporal data by analyzing the In-Phase (I) and Quadrature (Q) components of digital radio transmissions. The analysis of IQ data located at the lower end of the network stack relies on a large cache for data processing. In~\cite{Rajendran2019UnsupervisedFeatures}, an adversarial auto-encoder-based spectrum anomaly detector is proposed that uses features including power spectral density, signal bandwidth, and center frequency. 
While anomaly-based IDS has addressed the limitation of detecting unforeseen attacks in APD, there lacks a synthesized understanding of the cross-layer response of the signal under test (SUT) at various attack and upper layer domain knowledge, which limits its accuracy and efficiency. Meanwhile, evolved APD approaches utilizing the availability of software-defined 5G testbeds, large computational power, and advanced machine learning algorithms provide a rich prior-knowledge and deep understanding of the SUT and of the radio environment~\cite{Shao2020ConvolutionalSamples}, 
\cite{Mughal2018SignalClassifier}, \cite{Wang2021DevelopmentResearch}, \cite{Wang2021AI-PoweredOptimization}. 
To address these gaps in both IDS and APD, in this study, an anonymous jamming detection model for 5G NSA (Non-Standalone) is design and implemented. The contributions of this paper are summarized below:
\begin{itemize}
    \item A general approach of evaluating 5G NSA quality with the presence of WiFi-type interference in Long-Term Evolution (LTE) and New Radio (NR) cells is provided.
    
    \item Both supervised learning-based, and unsupervised learning-based algorithms are explored to construct an effective jamming detection model. 
    
    \item A 5G platform including user equipment, base station, core network and cyber attack generator are developed for data generation and proof of concept.  
    
    \item A Bayesian Network Model (BNM) is constructed for revealing causation and root, direct, and indirect causing of the jamming effects on the performance, thereby providing transparency to machine learning models.

\end{itemize}


    
    

The remainder of this paper is organized as follows: Sec.~\ref{jamming problem} formulates the problem, followed by a system description in Sec.~\ref{system description} and Sec.~\ref{causal}. Performance evaluation, conclusion and future work are provided in Sec.~\ref{evaluation} and Sec.~\ref{conclusion}.

\section{PROBLEM FORMULATION}\label{jamming problem}

\begin{figure}[htbp]
\centering
\includegraphics[width=0.46\textwidth]{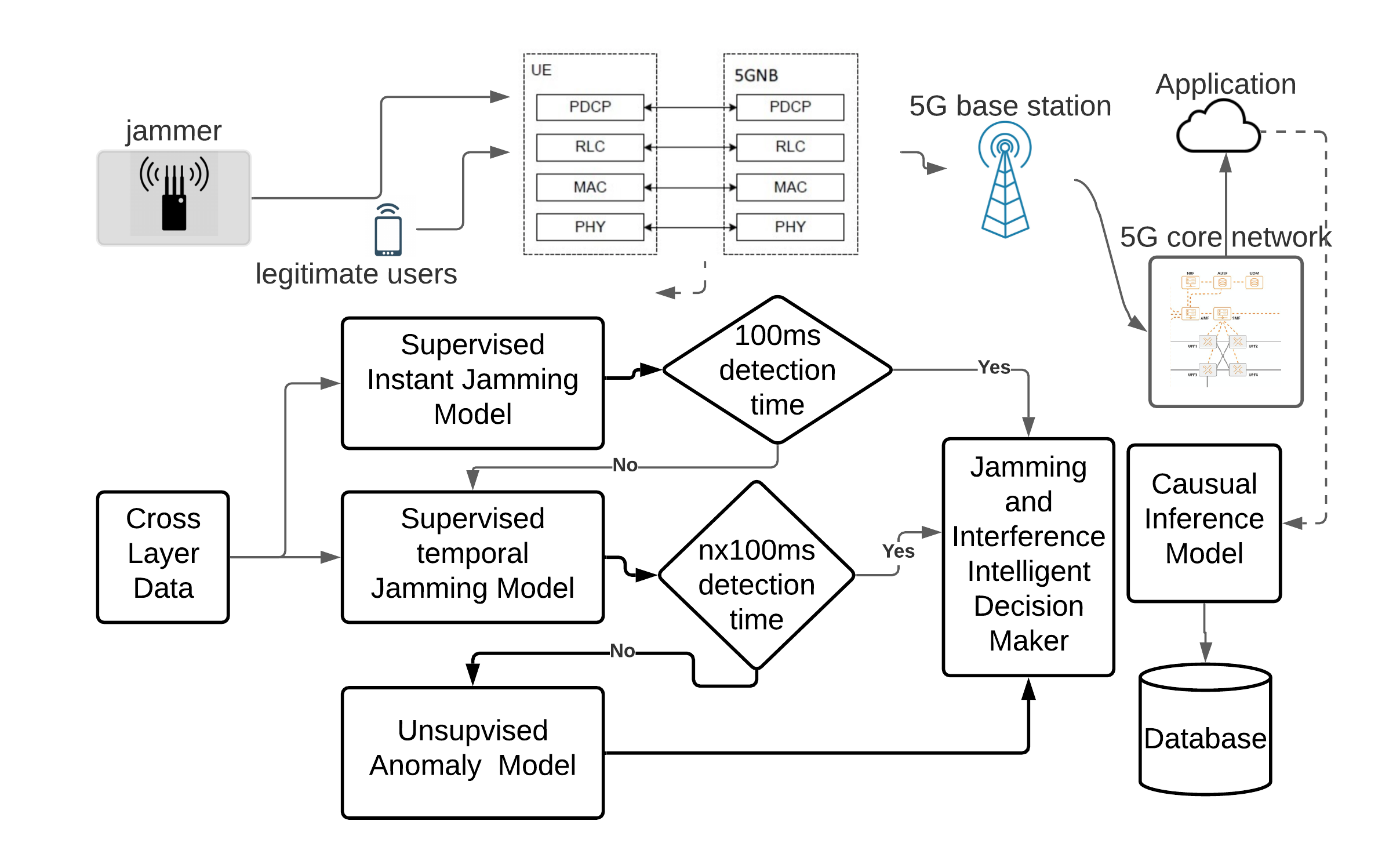}
\caption{System Overview}
\label{system}
\end{figure}

A jamming detection and cyber monitoring application of a 5G NSA wireless network as shown in  Fig.~\ref{system} was designed and implemented. The steps throughout the life cycle of jamming generation, communication configuration, data collection, intelligent analysis, and real-time feedback are integrated. With the presence of the jamming signal, communications with legitimate users would be impacted. Statistical information from cross-layer data plane including Physical Layer (PHY), Medium Access Layer (MAC), Radio Link Control (RLC), and Packet Data Convergence Control (PDCP) are collected and analyzed at the jamming detector module. The cross-layer data and side information available from the network and application layers are also input to the BNM for cyber inference analysis. The jamming detector and cyber inference analyzer module can be located at the base station (BS) to monitor the statistical information. It can also be a separate node fetching this information from the base station periodically. The former design, i.e., co-located at the BS, complies with the Open Radio Access Network (O-RAN)~\cite{O-RANAlliance2018O-RAN:RAN} architecture. In the latter setting, the jamming detector and analyzer module is located in a separate hardware unit and have minimal impact on the BS performance. 

Various types of jamming signals are generated and the corresponding statistical data is collected and stored in the implementation platform for model training and evaluation. The jamming discrimination can be formulated as two hypotheses:
\begin{equation*}
H_0: r[n] = s[n] + w[n], n = 1, ..., N\label{eq0}
\end{equation*}
\begin{equation*}
H_1: r[n] = s[n] + w[n] + j_m[n], n = 1, ..., N, \\
m\in[0, ..., M] \label{eq1}
\end{equation*}
where, $N$ is the total number of data samples collected, $M$ is the total number of different jamming signal types (including power level and center frequency changes), $r[n] = [r_{ue}[n], r_{bs}[n]]$  is  the concatenated list of received signals from both the User Equipment (UE) and the BS, $j[n]$  represents the concatenated unexpected signal generated by either interference or malicious jammers on both UE and the BS ends, $s[n]$ is the concatenated desired signal, $w[n]$ is the concatenated additive channel environment noise, $n$ is the index of discrete time slots, $m$ is the index of jamming type, where $m=0$ if the jamming type is unknown. 
Additionally,  
\begin{equation}\label{model input}
y[n] = f(r[n],r[n-1], ...,r[n-k]),
\end{equation}
where $y[n]$ is the statistical information from the BS and the UE for the communication link between them being observed for the past $k$ times slots.  

Our approach of jamming detection is designing a test statistic $\Lambda(y)$ and comparing it with a predefined threshold $\eta$, which can be denoted as:
$H_1 = \text{True}$ if $\Lambda(y) > \eta$ and  
$H_0 = \text{True}$ if $\Lambda(y) \leq \eta$.
The test statistic $\Lambda(y)$ and the threshold $\eta$ are determined by the selected machine learning models. Two types of models are proposed and compared: supervised discriminative models and unsupervised anomaly detection models. In the discriminative model, the objective is to maximize $P(\Lambda(y) > \eta|H_1)$ and maximize $P(\Lambda(y) < \eta|H_0)$ . In the unsupervised anomaly detection model, the objective is to maximize $P((\Lambda(y) > \eta, H_0)\cap (\Lambda(y) < \eta, H_1))$. 



\section{SYSTEM DESCRIPTION}\label{system description}
\subsection{Hardware Platform}\label{platform}

\begin{figure}[htbp]
\includegraphics[width=0.475\textwidth]{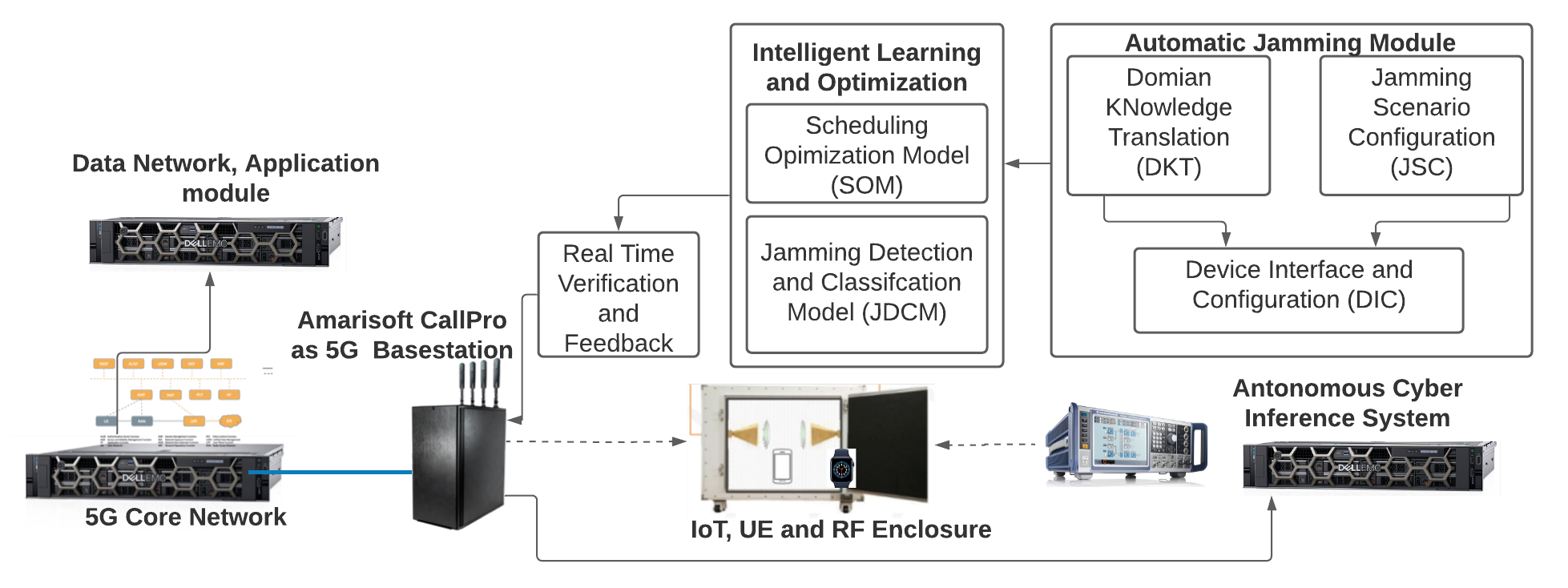}
\caption{System Hardware Platform Setup}
\label{hardware}
\end{figure}

The hardware platform used in our implementation of the 5G NSA testbed for experimental data collection is depicted in Fig.~\ref{hardware},
and consists of: User Equipment (UE), Base Station (BS), Core Network (CN), Cyber Attack Controller (CAC), Jamming Attack Radio Generator (JARG), and a Jamming Detection and Cyber Inference Model (JDCIM). The CAC generates different types of controlled jamming for training purposes. The JARG module emulates the jamming attacks on the legitimate communication link between UE and BS inside the Radio Frequency (RF) enclosure. The connections between BS and CN, CAC, and JARG are via an ethernet cable. Detailed information about this setup can be found in~\cite{Wang2021DevelopmentResearch}, \cite{Wang2021AI-PoweredOptimization}. 

The jamming waveforms are originally generated in MATLAB using the Signal Generation toolbox\texttrademark ~and  converted to Application Resource Bundle (ARB) files compatible with the R\&S SMW2000A Signal Generator, which provides a systematic way to generate PHY signal wave- forms for different wireless technologies. Table~\ref{waveform} shows the waveforms transmitted at the desired RF center frequency.

\begin{table}[htbp]
\caption{Jamming Waveforms Description}
\begin{center}
\begin{tabular}{|c|c|c|c|}
\hline
\textbf{Type} & \textbf{BW}& \textbf{Center Freq.}& \textbf{Power Levels (dBm)} \\
\hline
WiFi & 80 MHz & 2140 MHz & 0, -5, -11, -12, -13  \\
\hline
WiFi & 80 MHz & 1950 MHz & 0, -5, -11, -12, -13\\
\hline
WiFi & 80 MHz & 3490 MHz & -11, -12, -13 \\
\hline
\end{tabular}
\label{waveform}
\end{center}
\end{table}

The legitimate communication type selected is 5G NSA with NR cell Time Division Duplex (TDD) in band $N78$ and LTE cell Frequency Division Duplex (FDD) in band $B1$. The Center Frequency (CF) for the jamming signal is selected at the CF for $N78$ ($3.49$ GHz), uplink $B1$ ($1.95$ GHz), and downlink $B1$ ($2.14$ GHz). Different jamming power levels are chosen to compare their impacts. At these levels, the legitimate communication links are still affected but without a call drop, which is the area of interest for this study. 

Three types of machine learning models were constructed and integrated for comprehensive use case coverage. An instantaneous learning-based discriminative model is used for detecting known types of interference with short intervals. A temporal-based neural network model for detecting known types of interference is used to exploit underlying temporal correlations. An unsupervised learning-based anomaly detection model is used for detecting unknown types of jamming. The computational complexity and required number of data samples for training and evaluation of these three types of models increase incrementally. The performance of each type is evaluated using the same dataset generated as described in Table.~\ref{waveform}, and is detailed in Sec.~\ref{evaluation}. 

\subsection{Supervised Discriminative Model}

The designed jamming detection module for known jamming types feeds $y(n)$ in Eq.~\eqref{model input} as the input into a discriminative predictor. 
A single time-stamp (instantaneous) classification model needs shorter observation time in the on-field application, whereas a temporal-based model reveals underlying temporal correlations.
We have considered the performance for both instantaneous classification and temporal-based learning. Instantaneous classification is built with a single time stamp of data, which has the advantage of fast detection time within one sample period. In our implementation, the duration between consecutive time samples is 180 ms, which can be further reduced by higher performance hardware. Temporal-based learning is based on $k$ consecutive time steps, and thus provides $k$ adjacent data samples as the input for detection. 
In this work, we employ the LSTM as a temporal-based learning model, which has a chain structure comprising a specific neural network cell structure. 
The cell consists of three gates: input gate, output gate, and forget gate. The LSTM has a higher computational complexity for training and requires more data samples. In the training stage, the jamming information shown in Table~\ref{waveform} is automatically recorded and labeled in a local database. The independent variables used as input features to both the instantaneous classification model and the temporal-based model are downlink/uplink bitrate, downlink/uplink packet rate, downlink/uplink retransmission rate, Physical Uplink Shared Channel Signal Quantity and Signal to Interface and Noise Ratio (PUSCH SNR), Channel Quality Indicator (CQI), power headroom, energy per resource element, uplink path loss, downlink/uplink Modulation and Coding Scheme (MCS), and the average turbo decoder rate.



The discriminative models are trained using positive (known target jamming types) and negative cases (no jamming present).
However, when the jamming types are unknown, an unsupervised anomaly detection approach is adopted with the model trained only with standard signals (without jamming present). 
The relationship of the supervised discriminative models and unsupervised anomaly detection model and their usage in our proposed system is shown in Fig.~\ref{system}.


\subsection{Unsupervised Model} \label{Unsupervised}
To identify jamming attacks without prior training on a particular jamming type, a network intrusion detection system is proposed in~\cite{Mirsky2018Kitsune:Detection}, which is trained only on the normal network traffic and then detects attacks as anomalies. We have utilized the auto-encoder based architecture proposed in~\cite{Mirsky2018Kitsune:Detection} to detect previously unseen jamming attacks.

\begin{figure}[htbp]
    \centering
	\includegraphics[width=0.475\textwidth]{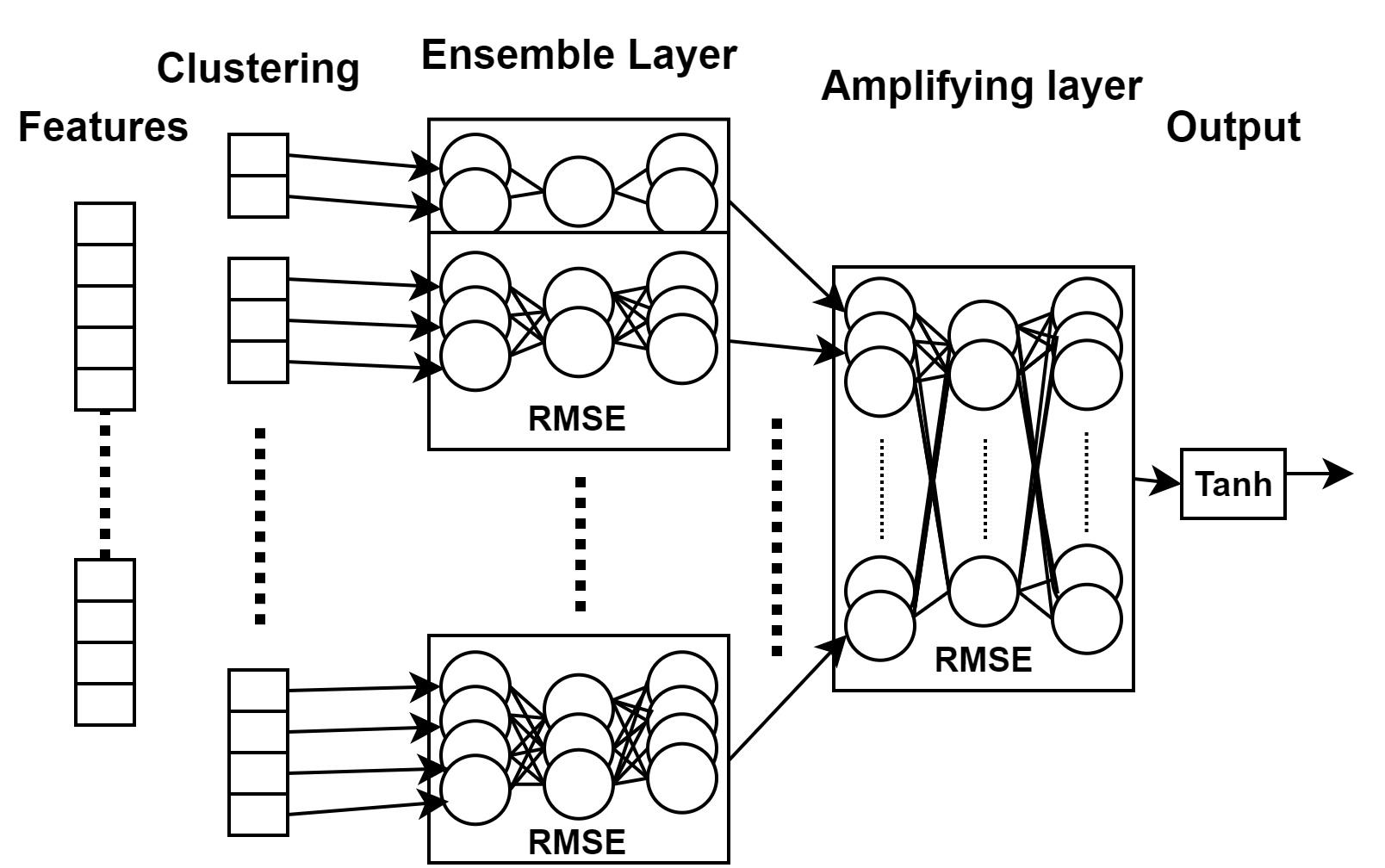}
	\caption{The ensemble of auto-encoders, reproduced from \cite{Mirsky2018Kitsune:Detection}. }
	\label{ensemble-autoencoder}
\end{figure}

Auto-encoders are trained to recreate the input data. As long as new data follows the same distribution in the feature space as the data it has been previously trained on, the auto-encoder can recreate them. Working under the hypothesis that a jamming attack is observed in the feature space, we can expect that the auto-encoder will fail to successfully recreate it and generate a large error that can be detected. The training data comprising only non-interference samples is parsed using the input features described in the previous section. 
These features are clustered hierarchically and each cluster is passed to an auto-encoder. This gives an ensemble of $d$ auto-encoders, where $d$ is the number of clusters. The root mean squared errors from these auto-encoders are concatenated and passed to another auto-encoder. This secondary auto-encoder amplifies the errors from the ensemble. The overall architecture is described in Fig.~\ref{ensemble-autoencoder}. The output from this auto-encoder, in the range $[0,\infty)$, is bound to the range $[0,1]$ by passing it through the hyperbolic tangent function to get the output score denoting the probability of jamming presence. 


\section{Causality Analysis}\label{causal}
\begin{figure}[htbp]
\includegraphics[width=0.475\textwidth]{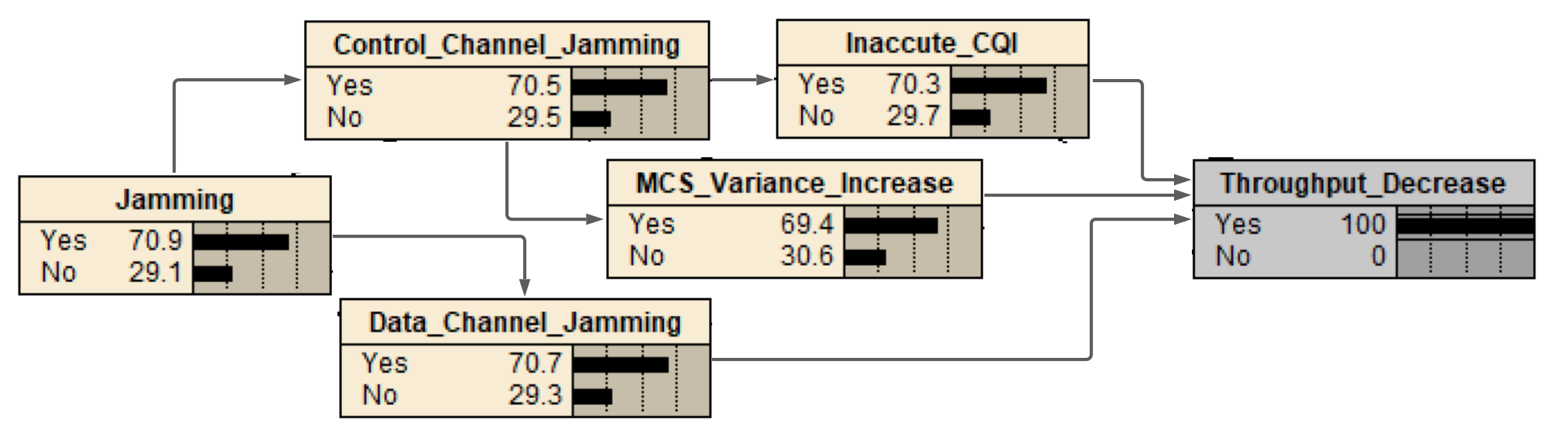}
\caption{BNM Based Inference for Throughput Degradation}
\label{causal_topology}
\end{figure}

A BNM is a Directed Acyclic Graph (DAG) that discovers and represents dependencies among random variables from observational data~\cite{Kyburg1991ProbabilisticInference.}. The objective of the proposed causal inference analysis is to enhance the accuracy of predicting jamming presence and assess its effect on significant quality attributes in various use cases. Due to a partial loss of information in transition from empirical information to conditional probability tables,  methods of probabilistic inference from learning data may have shortcomings such as high computational complexity and the cumulative error increasing exponentially as the number of nodes in the DAG increases  \cite{Terentyev2007METHODNETWORKS}. To address the computational complexity, in the proposed method, a topology is constructed using both the domain knowledge in wireless networks and reverse engineering of the data in the jamming detection study. Fig.~\ref{causal_topology} shows indirect, direct, and root cause for the throughput performance through domain knowledge. In DAG-based topology, the relationship and probability from node `jamming', `Inaccurate CQI', `MCS Variance Increase', `Data Channel Jamming', and the impact to `Throughput Decrease' are established by domain knowledge and verified by empirical data. 

When physical layer jamming is present in the control channel (CCH) dedicated to control information between the UE and the network, the reactions of CQI and MCS are different compared to when the jamming is present in the data channel (DCH) dedicated for data information. As demonstrated in~\cite{Wang2021AI-PoweredOptimization}, CCH jamming causes the incorrect CQI reported by UE, and further causes the increased variance in the MCS distribution, which does not exist in DCH jamming. Through domain knowledge of the 5G protocol, we know that the MCS value or the range of MCS values are set automatically based on the value of CQI, which establishes the causality relationship between MCS and CQI in Fig.~\ref{causal_topology}. The RF environment and the presence of jamming affect the CQI values reported by UE. The throughput is determined by factors from multiple layers, however, in this study, we prioritize the factors from the physical layer and keep the other factors constant to simplify the structure. Eq.~\eqref{bnm} is used to calculate the probability of sentinel event $S$, given a set of different unobserved ($C_u$) and observed causes ($C_i$)s to simply the casual relationship. Furthermore, the BNM is  used to calculate the probability of observing a cause, in our case, the root cause $r$, given that an effect (e.g., throughput decrease) has occurred due to the observed causes.  
\begin{equation*}
\begin{split}
& P(S|r,C_1,\cdots,C_5, C_{u_1},...,C_{u_n}) =\\ &\sum_{C_u}{P(S|r,C_1,...,C_5)P(C_{u_1},...,C_{u_n})}= P(S|r,C_1,...,C_5)\\
\end{split}
\end{equation*}




\section{Performance Evaluation}\label{evaluation}
\subsection{Supervised Jamming Detection Performance}

The performance evaluation of the proposed machine learning-based jamming detection models is based on data collected as per Table~\ref{waveform}. 
For instantaneous classification, a number of different models are used to compare the performance, namely logistic regression, Gaussian Naive Bayes, $K$-neighbors classifier, decision tree, and random forest. A train-test split of $75\%$-$25\%$ is used in these evaluations. The accuracy of distinguishing between the cases of jamming present and no jamming present is significantly high with tree-based classification. When detecting the listed jamming types from the no jamming scenarios, the random forest algorithm gives $100\%$ accuracy for all jamming types. The reason behind the superior performance using decision trees is that decision trees or random forest models divide the feature space into smaller and smaller regions, whereas other methods such as logistic regression fit a single line to divide the space exactly into two. The light-weighted tree based classifier enables the possibility of migrating the system to UE or Internet of Things (IoT) devices. 



\begin{figure}[htbp]
\includegraphics[width=0.475\textwidth]{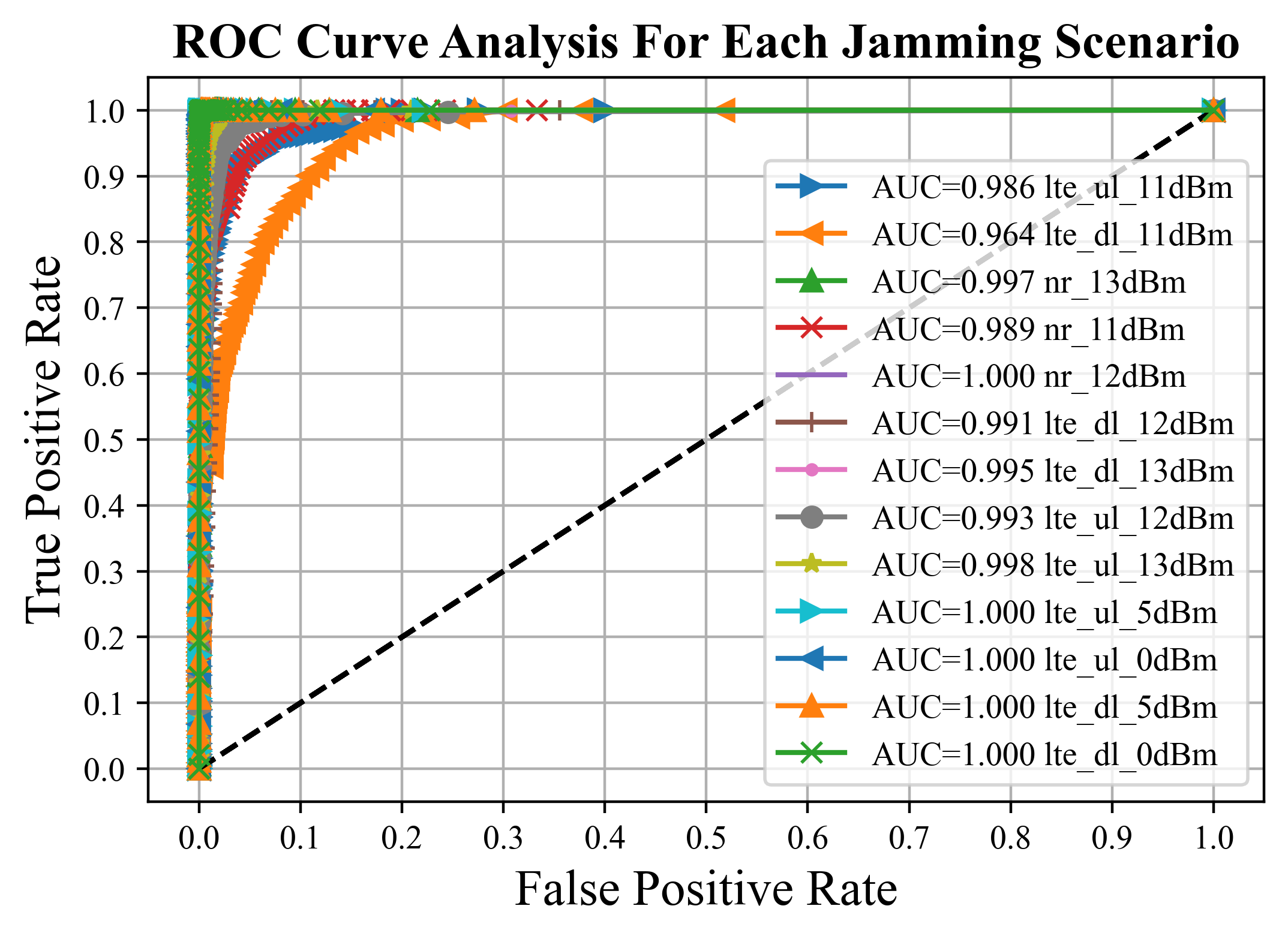}
\caption{The Overview of AUC for Each Jamming Type}
\label{roc_everything}
\end{figure}

In jamming classification for differentiating types of jamming, tree-based classification shows slightly lower but still significantly high performance. As an overview of the jamming classification methods, the AUC values to detect any specific jamming type against all other scenarios are shown in Fig.~\ref{roc_everything}.



In order to exploit the correlation amongst the adjacent time stamps of data, a temporal-based learning scheme based on the LSTM is also implemented. In our implementation, we use $k=2$ consecutive data samples, one belonging to the LTE cell and the other sample from the 5G NR cell. Our LSTM implementation uses a single hidden layer consisting of $100$ LSTM units, followed by a fully-connected layer with $100$ nodes, with a final output layer with $2$ nodes for binary jamming detection. The precision-recall performance for the temporal-based jamming detector is summarized in Table~\ref{lstm_performance}. 

\begin{figure}[htbp]
\centering
\captionsetup{justification=centering}
\includegraphics[width=0.475\textwidth]{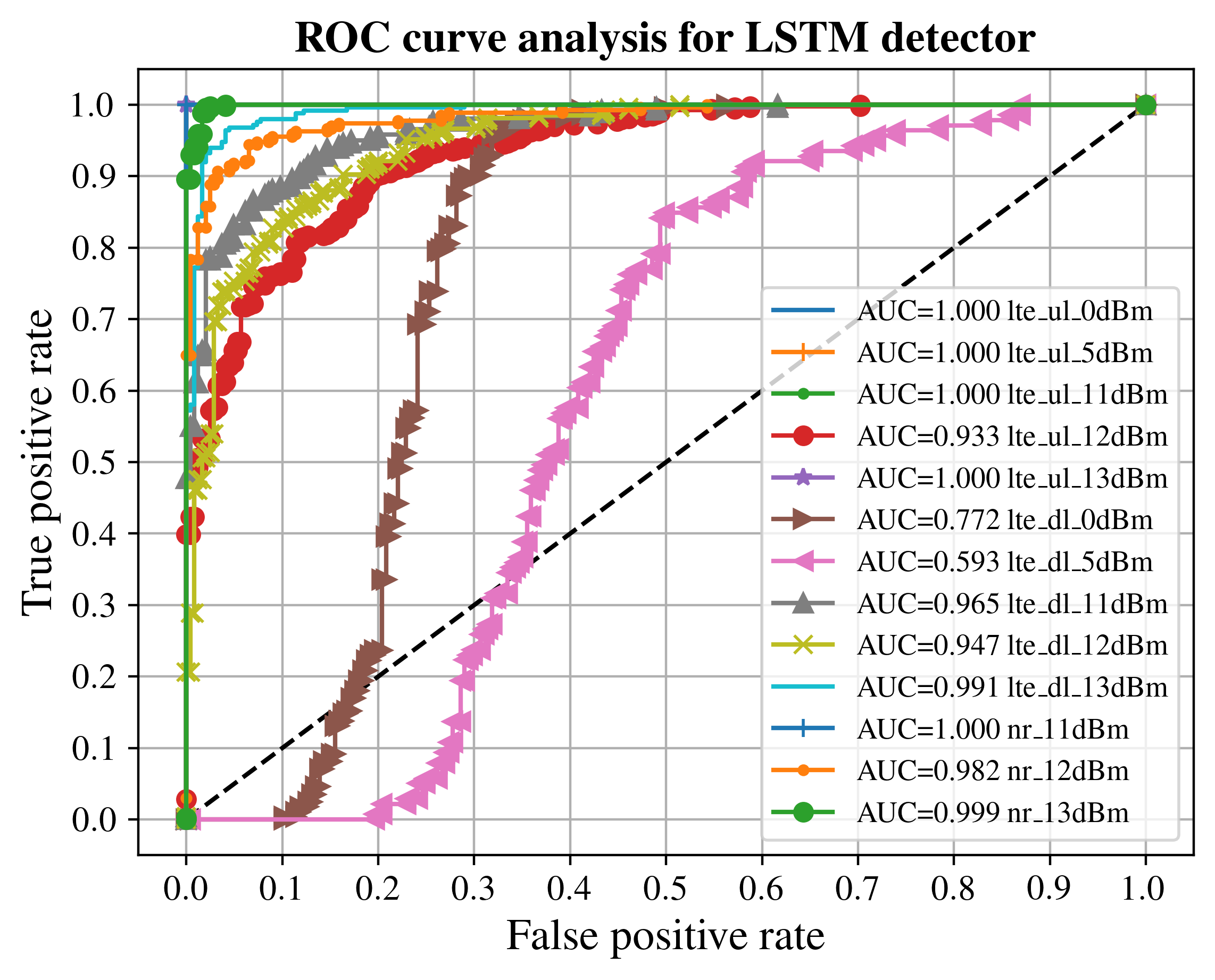}
\caption{AUC for each jamming type LSTM-based detection}
\label{roc_lstm}
\end{figure}

\begin{table}[htbp]
\centering
\captionsetup{justification=centering}
\caption{Temporal-based (LSTM) Detector Performance}
\begin{center}
\begin{tabular}{|c|c|c|c|c|c|}
\hline
\textbf{}                         & \multicolumn{5}{c|}{$k=2$ time steps}                                                                           \\ \hline
\textbf{Center frequency (GHz)}   & \multicolumn{2}{c|}{\textbf{1.95}} & \multicolumn{2}{c|}{\textbf{2.14}} & \textbf{3.49}                     \\ \hline
\textbf{Interference power (dBm)} & \textbf{0}      & \textbf{-5}      & \textbf{0}      & \textbf{-5}      & \multicolumn{1}{l|}{\textbf{-11}} \\ \hline
\textbf{Avg. accuracy (\%)}       & 99.7            & 95.9             & 84.4            & 59.3             & 99.8                              \\ \hline
\textbf{Avg. precision (\%)}      & 99.8            & 95.2             & 78.9            & 46.7             & 100                               \\ \hline
\textbf{Avg. recall (\%)}         & 99.5            & 97.2             & 96.7            & 85.8             & 99.7                              \\ \hline
\end{tabular}
\label{lstm_performance}
\end{center}
\end{table}

We show the performance of the LSTM jamming detector in different jamming scenarios as ROC curves in Fig.~\ref{roc_lstm}, with the AUC as a single performance metric to compare with the instantaneous classification models.

For LTE Downlink (DL) band (2140 MHz) jamming at $-5$ dBm, the performance is lower due to the limited training data sample size. The BNM approach from Sec.~\ref{causal} is used for performance enhancement. When applying the model in Fig. \ref{causal_topology} to the LTE DL $-5$ dBm, the training data shows with presence of control channel (CCH) jamming, the probability of a MCS variance increase detected is $0.845$ and inaccurate CQI detected is $0.503$. Thus a MCS variance increase being observed, the jamming recall increases to 86.4\% and precision to 80.9\%.   

\subsection{Unsupervised Jamming Detection Performance} \label{evaluation_Unsupervised}

\begin{figure}[htbp]
    \centering
	\includegraphics[width=0.475\textwidth]{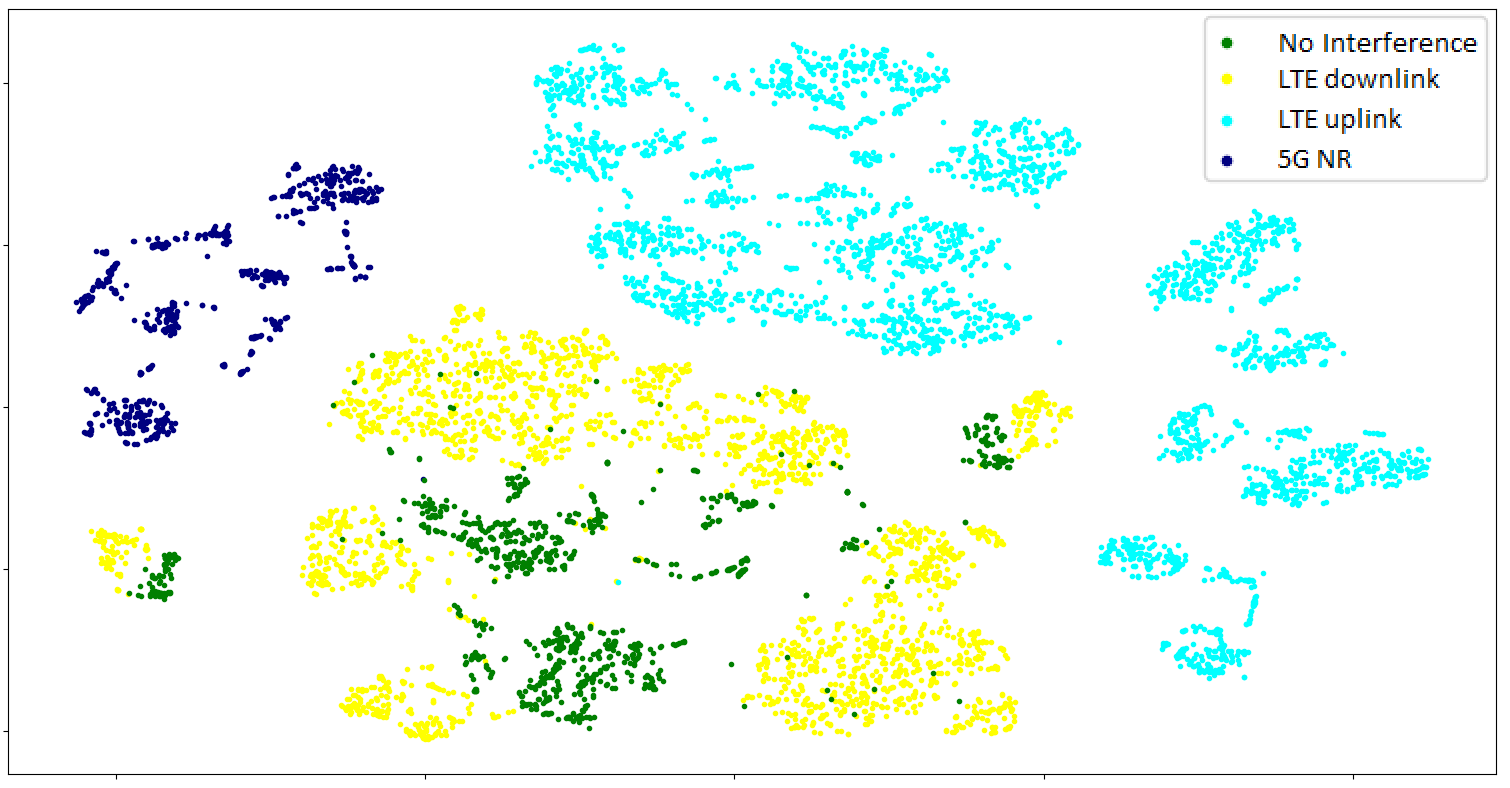}
	\caption{The t-SNE projection of the feature space for the interference data}
	\label{TSNE}
\end{figure}

To evaluate the performance of the approach described in Sec.~\ref{Unsupervised}, the model was trained on $596$ samples of non-interference data. The remaining non-interference data, and samples from several unknown types of jamming, 5G new radio, LTE uplink, with multiple power levels for the LTE interference, were concatenated, totaling to $4125$ samples, and used to test the trained model. The proposed model performs remarkably well in detecting the unknown jamming attacks. The per class accuracy are summarized in table \ref{AE_performance}.  
The approach however fails to detect interference in LTE DL bands. The reason for this becomes apparent by looking at a t-SNE projection of the entire data set. \mbox{Figs.~\ref{TSNE}} demonstrates how the LTE uplink and NR interference are separable from the no interference data. But the projection shows that the features for LTE DL and no interference have overlapping distributions and the proposed unsupervised approach lacks the information to distinguish between the two.

\begin{table}[htbp]
	\centering
	\captionsetup{justification=centering}
	\caption{Ensemble auto-encoder Performance}
	\begin{center}
		\begin{tabular}{|c|c|c|c|c|c|}
			\hline
			\multirow{2}{*}{} &      \textbf{No }      & \textbf{5G NR }  &      \multicolumn{3}{c|}{\textbf{LTE uplink}}      \\ \cline{3-6}
			                  & \textbf{ interference} & \textbf{-11 dBm} & \textbf{0 dBm} & \textbf{-5 dBm} & \textbf{-11dBm} \\ \hline
			\textbf{Accuracy} &          96.7          &       100        &      100       &       100       &       100       \\ \hline
		\end{tabular}
		\label{AE_performance}
	\end{center}
\end{table}


\begin{figure}[htbp]
	\includegraphics[width=0.475\textwidth]{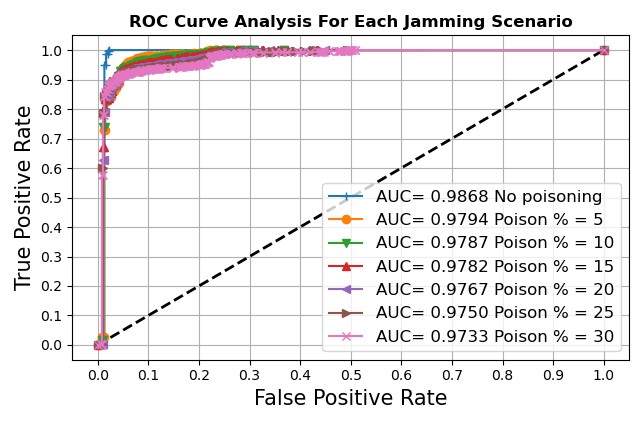}
	\caption{AUC for unknown jamming with noisy training data}
	\label{roc_unsupervized}
\end{figure}

\begin{figure}[htbp]
    \centering
	\includegraphics[width=0.475\textwidth]{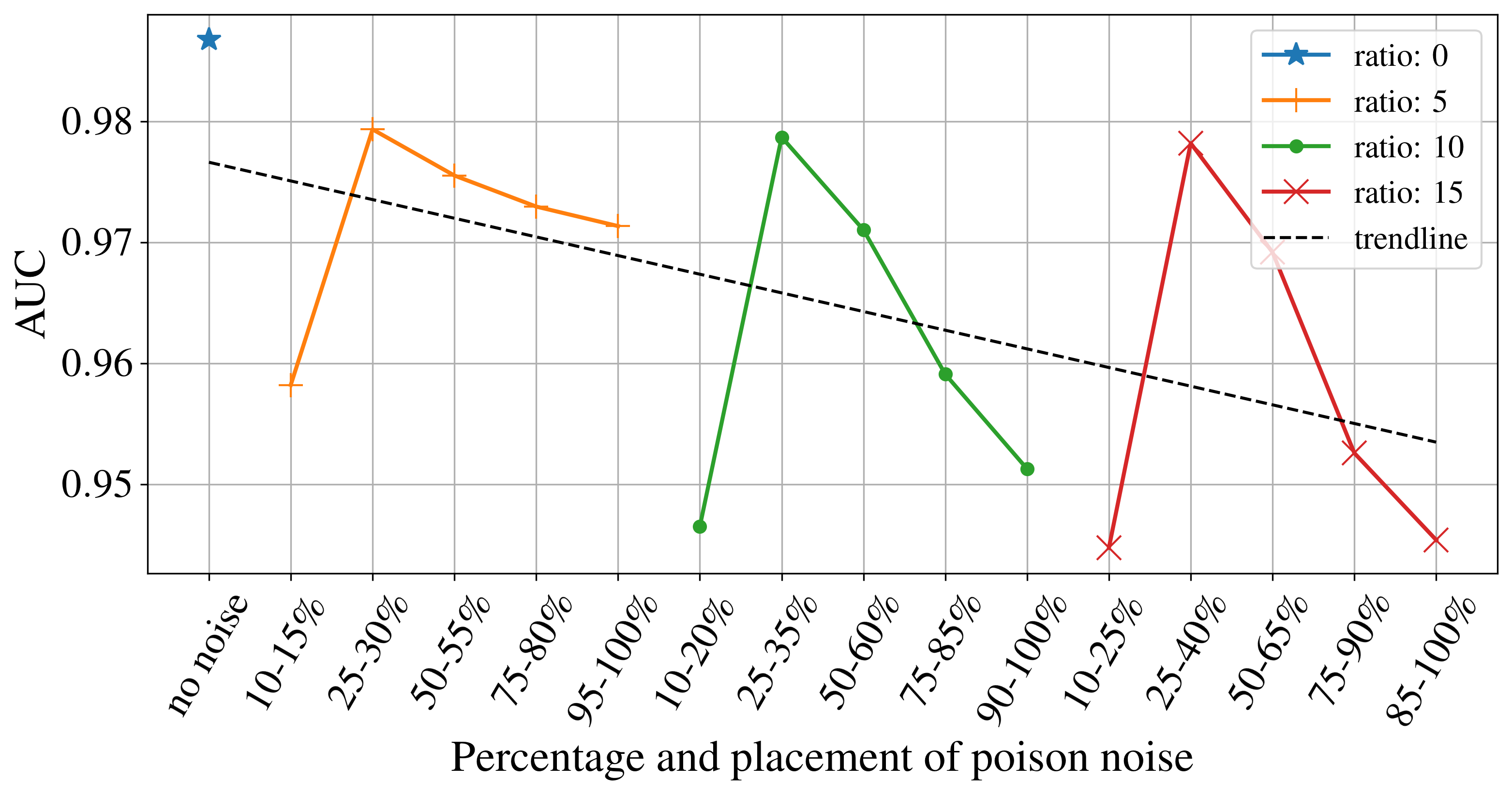}
	\caption{Temporal effect of the adversarial training }
	\label{AUC_aggregrate}
\end{figure}

Adversarial training is one of the known vulnerabilities of learning-based approaches. Poisoning the training data with adversarial examples can severely compromise the capability of a learning based model.  The effect of poisoning of the training data with jamming samples was evaluated. The first $10\%$ phase of the training, during which the feature clustering algorithm runs, is completely resistant to poisoned training data. Beyond that the model can withstand a significant percentage of poisoned training samples with graceful gradual degradation.
Fig.~\ref{roc_unsupervized} demonstrates the results when differing percentage of noise is introduced at the beginning of the training, right after the clustering. In addition, the stage at which the samples are poisoned shows significant impact. 
The model performance was tested with different fractions of poisoned training data placed at different stages during the training. The AUC performance in this setting is summarized in Fig.~\ref{AUC_aggregrate}. As previously stated, the first $10\%$ of the training is immune to poisoning. Beyond that, it is empirically observed that early and late training is significantly more vulnerable compared to the mid-training period. Overall, the proposed model is not only remarkably effective in detecting unknown jamming types, but is also robust against adversarial training.    




\section{Conclusion}\label{conclusion}
In this work, an anonymous jamming detection and classification scheme is introduced, which is based on monitoring and analysis of cross-layer statistical parameters. In particular, supervised learning based instantaneous classification and temporal-based models are developed along with an unsupervised learning-based anomaly detection approach, which, as separate parts of a combined approach, achieve both high accuracy and robust detection for known and unknown jamming types respectively. A causality analysis approach using BNM is developed to identify the root cause of Key Performance Indictor (KPI) degradation caused by jamming, thus adding transparency to the model. Development of the temporal-based supervised learning jamming detection scheme as well as the extension of the Bayesian inference scheme for a deeper root cause analysis is part of future work.

\section*{Acknowledgment}
This work was made possible through the collaboration
between Dr. Jeff Reed, the Willis G. Worcester professor
in Bradley Department of Electrical and Computer Engineering, and Dr. Laura Freeman, Director 
of the Hume Center for National Security and Technology’s 
Intelligent Systems Lab in Virginia Tech, funded by Deloitte \& Touche LLP’s Cyber Strategic Growth Offering led by Deborah Golden.

\bibliographystyle{IEEEtran} 
\bibliography{references}

\vspace{12pt}

\end{document}